\documentclass[letterpaper, 10 pt, conference]{ieeeconf}  % Comment this line out if you need a4paper

\IEEEoverridecommandlockouts                              % This command is only needed if 
\usepackage[numbers]{natbib}
\usepackage{graphicx}
\usepackage{amsmath}
\usepackage{hyperref}
\usepackage{algpseudocode}
\usepackage{algorithm}
\usepackage{multirow}
\usepackage[normalem]{ulem}
\usepackage{booktabs}
\usepackage{lipsum}

\usepackage[utf8]{inputenc} % allow utf-8 input
\usepackage[T1]{fontenc}    % use 8-bit T1 fonts
\usepackage{url}            % simple URL typesetting
\usepackage{amsfonts}       % blackboard math symbols
\usepackage{nicefrac}       % compact symbols for 1/2, etc.
\usepackage{microtype}      % microtypography
\usepackage{xcolor}         % colors
\usepackage{amssymb}         % colors
\usepackage{color}
\usepackage{blindtext}
\let\labelindent\relax
\usepackage{enumitem}
\usepackage{arydshln}
\usepackage{wrapfig}
\usepackage{bbm}
\usepackage{breqn}
\usepackage{subcaption}

\newcommand{\method}{\textsc{P3-PO}}
\newcommand{\italmethod}{\textit{P3-PO}}
\newcommand{\website}{\href{https://point-priors.github.io/}{point-priors.github.io}}

\hypersetup{
    colorlinks=true,
    linkcolor=blue,
    filecolor=magenta,      
    citecolor=blue,
    urlcolor=blue,
}

\newcommand{\xxnote}[3]{}
\ifx\hidenotes\undefined
  \renewcommand{\xxnote}[3]{\color{#2}{#1: #3}}
\fi

\title{\LARGE \bf
\method{}: Prescriptive Point Priors for Visuo-Spatial \\Generalization of Robot Policies
}

\author{
Mara Levy$^{1}$\thanks{Correspondence to: mlevy@umd.edu} \qquad Siddhant Haldar$^{2}$ \qquad Lerrel Pinto$^{2}$ \qquad Abhinav Shirivastava$^{1}$ \\[0.5em]
$^{1}$University of Maryland \qquad $^{2}$New York University\\[0.5em]
{\small \tt \website{}}
}

\begin{document}

\maketitle
\thispagestyle{empty}
\pagestyle{empty}

%%%%%%%%%%%%%%%%%%%%%%%%%%%%%%%%%%%%%%%%%%%%%%%%%%%%%%%%%%%%%%%%%%%%%%%%%%%%%%%%
\begin{abstract}
Developing generalizable robot policies that can robustly handle varied environmental conditions and object instances remains a fundamental challenge in robot learning. While considerable efforts have focused on collecting large robot datasets and developing policy architectures to learn from such data, na\"ively learning from visual inputs often results in brittle policies that fail to transfer beyond the training data. This work presents Prescriptive Point Priors for Policies or \method{}, a novel framework that constructs a unique state representation of the environment leveraging recent advances in computer vision and robot learning to achieve improved out-of-distribution generalization for robot manipulation. This representation is obtained through two steps. First, a human annotator prescribes a set of semantically meaningful points on a single demonstration frame. These points are then propagated through the dataset using off-the-shelf vision models. The derived points serve as an input to state-of-the-art policy architectures for policy learning. Our experiments across four real-world tasks demonstrate an overall 43\% absolute improvement over prior methods when evaluated in identical settings as training. Further, \method{} exhibits 58\% and 80\% gains across tasks for new object instances and more cluttered environments respectively. Videos illustrating the robot's performance are best viewed at \website{}.

% Developing generalizable robot policies that can robustly handle varied environmental conditions and object instances remains a fundamental challenge in robot learning. While considerable efforts have focused on collecting large robot datasets and developing policy architectures to learn from such data, naively learning from visual inputs often results in brittle policies with limited generalization beyond the training data. This work presents Point Priors for Policies or \method{}, a novel framework that leverages recent advances in computer vision and robot learning to achieve improved out-of-distribution generalization for robot manipulation.
% \method{} is able to generalize by constructing a unique state representation for the environment. This representation is obtained through two steps. First a set of points is obtained from the visual input utilizing state-of-the-art computer vision models. Next these points are distilled into a graph representation which encodes inter-object spatial relationships. The graph is then combined with state-of-the-art policy architectures for policy learning. Experiments across four real-world tasks demonstrate an overall \SH{XX}\% absolute improvement over prior methods when evaluated in identical settings as training. Further, \method{} exhibits \SH{XX}\% and \SH{XX}\% gains across tasks for new object instances and more cluttered environments respectively. Videos illustrating the robot's performance are best viewed at \website{}.

\end{abstract}

%%%%%%%%%%%%%%%%%%%%%%%%%%%%%%%%%%%%%%%%%%%%%%%%%%%%%%%%%%%%%%%%%%%%%%%%%%%%%%%%
%%%%%%%%%%%%%%%%%%%%%%%%%%%%%%%%%%%%%%%%%%%%%%%%%%%%%%%%%%%%%%%%%%%%%%%%%%%%%%%%
\section{Introduction}
A long standing goal in robotics has been to develop robot policies that are robust to environmental changes and can operate across variations in spatial configurations and object instances. While significant advances have been made in this direction for computer vision~\cite{dalle3,imagen} and natural language processing~\cite{gpt4,gemini,llama}, the majority of robot policies remain confined to controlled laboratory environments with carefully designed settings. Robotic policies struggle to generalize to real-world scenarios because of the challenges and high costs associated with gathering diverse, high-quality robotic data. 
% This stands in contrast to fields like computer vision and natural language processing, which have succeeded in the wild largely due to the large, varied datasets available for model training.\LP{Last line is repetitive}

\begin{figure}[t]
    \centering
    \includegraphics[width=0.8\linewidth]{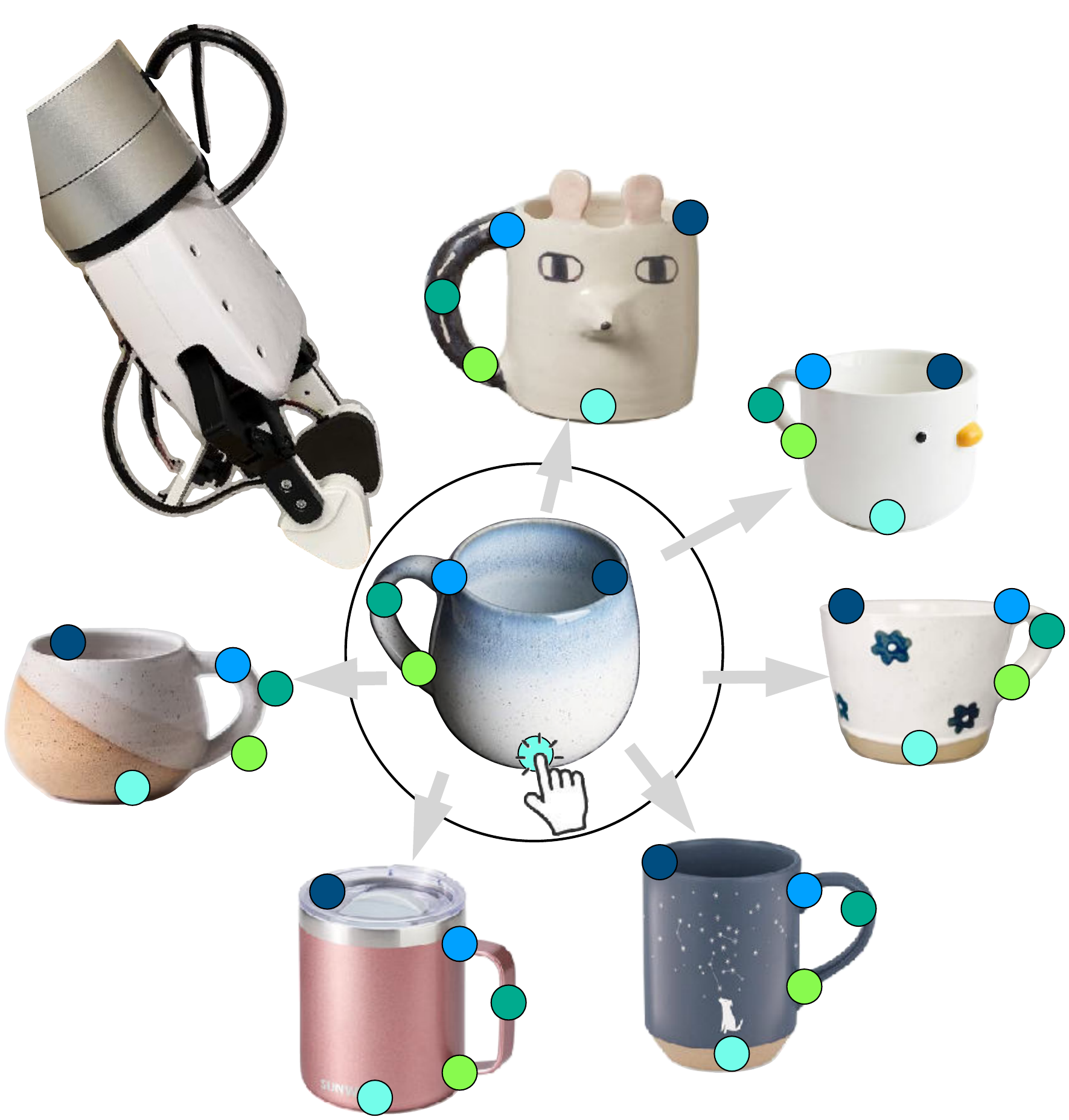}
    % \vspace{-.75em}
    \caption{A human prescribes key points one time for one instance of an object and those points are transferable to all other instances of the same object.}
    \vspace{-1.5em}
    \label{fig:intro}
\end{figure}

\begin{figure*}[t]
    \centering
    \includegraphics[width=\linewidth]{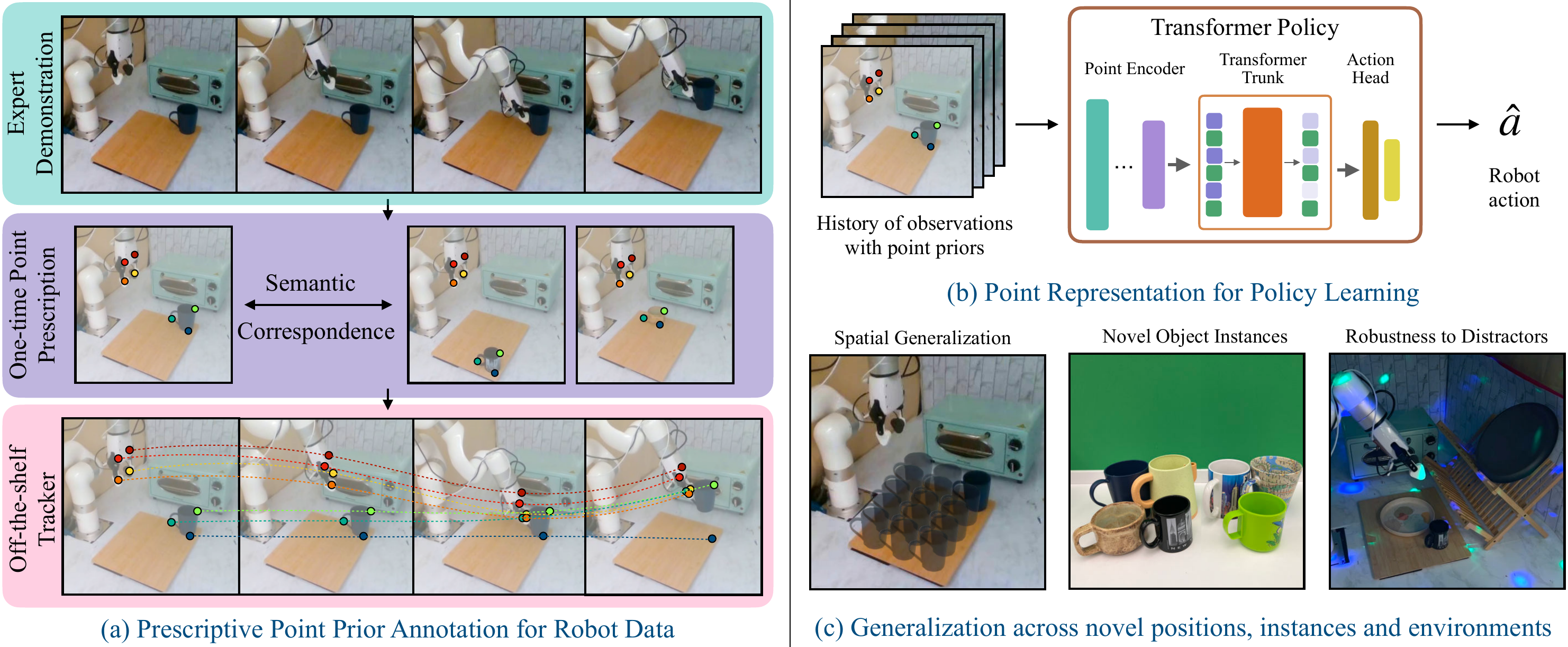}
    \vspace{-1em}
    % \caption{An overview of \method{}. (a) A human annotates key points of \textbf{just one frame} and semantic correspondence is able to identify the same points at the start of each following episode. (b) The final graph representation is formed using the key points from the previous step and either the ground truth depth or an estimated metric depth~\cite{depth_anything_v2}. The current graph representation along with the prior $t$ steps of history are passed into a transformer based policy which outputs the action. (c) \method{} is able to generalize to a wide range of spatial locations as well as new instances of each object. It is also robust to having other distracting objects in the scene. \SH{need to emphasize that point labeling is done just once - see second-last para in intro.}}
    \caption{Overview of the Prescriptive Point Priors for Policies (\method{}) framework. (a) A human annotator prescribes a set of semantically meaningful key points on a \textbf{single demonstration frame}, typically in under 5 seconds. Off-the-shelf computer vision models are then used to automatically propagate these key points throughout the entire dataset without further human input. (b) The derived key points are leveraged by a transformer policy to predict the action. (c) \method{} enables learning policies with improved generalization capabilities, including spatial generalization (i.e. generalization to new locations), generalization to novel object instances, and robustness to background distractors. \method{} combines the strengths of vision and policy prediction methods through simple yet effective human-prescribed semantic guidance.}
    \vspace{-1em}
    \label{fig:method}
\end{figure*}

Recent efforts aim to address this data problem by either aggregating existing robot datasets under a common framework~\cite{rtx} or collecting extensive real-world datasets through easy-to-use teleoperation tools~\cite{aloha,dobbe,umi,rum}. However, aggregated datasets suffer from inconsistencies across actions recorded for different projects~\cite{rtx}, while large teleoperated collections, though useful, are often specific to a single robot type~\cite{dobbe,aloha} and it is unclear whether these approaches would scale for different robot morphologies. Consequently, developing generalized robot models still largely depends on collecting more expert demonstrations. 

% \SH{why keypoints? Smoothen transition - why not use data priors, observational/representaion priors - SSL, detection, recognition?}

% \SH{alternate to large scale data is to look at stronger visual representations - sengmentaion (GROOT), pose estimators( David Held) - either lose too much info, requires apriori knowledge of class/model of objects. Is there a representation that is flexible, strong priors, object centric representation - keypoint -> no training per object but important spatial info}

% \SH{full 3D is probably too much}

One way to get around this data problem is to use strong representation priors that transfer across scenarios and feed these representations as input into existing policy architectures. While priors such as object proposals~\cite{sieb2020graph,viola,groot} and pose estimation~\cite{pan2023tax,vitiello2023one} have been used in prior work, they often lose information which makes policy learning harder or require accurate modeling of the object poses to make the policy work. In this work, we explore if there exists a representation that is flexible, serves as a strong prior, and can provide the object-centric abstraction of a scene to enhance generalization. Compared to segmentation and object models, a point-based representation retains fine-grained spatial information without requiring accurate modeling of object boundaries or poses. By representing objects and scenes as a set of unstructured points, these representations extract only the essential geometric relationships between relevant elements in the scene. This allows the policy to focus exclusively on the key spatial interactions. 

We present Prescriptive Point Priors for Policies or \method{}, a novel framework that leverages the generalization capabilities of state-of-the-art computer vision models alongside state-of-the-art robot policy architectures. Through \method{}, we demonstrate improved spatial generalization, the ability to generalize to novel object instances, and robustness to large environmental changes. \method{} is built on three key ideas. First, a human annotator prescribes a set of semantically meaningful points on a single demonstration frame, a process that often requires less than 5 seconds. Second, a diffusion-based visual correspondence model~\cite{dift} and a state-of-the-art point tracker~\cite{cotracker} are used to seamlessly transfer these points to the entire dataset without further human involvement. Third, the derived points are combined with state-of-the-art policy architectures like BAKU~\cite{baku} for learning robust robot policies. The novelty of \method{} lies in strategically combining the strengths of vision models and policy methods, yielding a simple yet effective approach for generalizable robot learning.

% it leverages a diffusion-based visual correspondence model~\cite{dift} and a state-of-the-art point tracker~\cite{cotracker} to reduce image-based visual inputs into a set of points on the image. Second, these points are converted into a graph-based representation that encodes inter-object distances in the scene, improving spatial generalization. Third, this graph-based representation is combined with state-of-the-art policy architectures like BAKU~\cite{baku} for learning robust robot policies. The novelty of \method{} lies in carefully combining the strengths of vision models and policy architectures which leads to a framework for generalizable robot learning. \LP{Maybe add somewhere that it is simple, requires humans clicking only once per task. Takes less than 5 seconds.}

We demonstrate the effectiveness of \method{} through experiments on four real-world tasks in an xArm Kitchen environment. Our main findings are summarized below:

\begin{enumerate}[leftmargin=*,align=left]
    \item \method{} exhibits an overall $43$\% improvement over prior state-of-the-art policy learning algorithms across 4 real world tasks (Section~\ref{subsec:spatial_gen}). 
    \item \method{} generalizes to novel object instances, exhibiting a $58$\% improvement on a set of held-out objects as compared to prior work (Section~\ref{subsec:novel_gen}).
    \item Policies trained with \method{} are robust to the presence of background distractors (Section~\ref{subsec:distractor}) and work with both true depth and predicted metric depth from state-of-the-art models like Depth Anything~\cite{depth_anything_v1,depth_anything_v2} (Section~\ref{subsec:da}).
\end{enumerate}

All of our datasets, and training and evaluation code are publicly available. Videos of our trained policies can be seen here: \website{}.

\section{Related Works}
\subsection{Imitation Learning (IL)} 
Imitation Learning (IL)~\cite{imitation_learning} refers to training policies with expert demonstrations, without requiring a predefined reward function. In the context of reinforcement learning (RL), this is often referred to as inverse RL~\cite{irl_1, irl_2}, where the reward function is derived from the demonstrations and used to train a policy~\cite{wayex, rot, fish, gail, nair2020awac}. While these methods reduce the need for extensive human demonstrations, they still suffer from significant sample inefficiency. As a result of this inefficiency in deploying RL policies in the real world, behavior cloning (BC)~\cite{Pomerleau-1989-15721, torabi2019recent, schaal1996learning, ross2011reduction} has become increasingly popular in robotics. Recent advances in BC have demonstrated success in learning policies for both long-horizon tasks~\cite{sequential_dexterity,learning_to_generalize,clipport} and multi-task scenarios~\cite{baku,roboagent,rtx,track2act}. However, most of these approaches rely on image-based representations~\cite{zhang2018deep,baku,diffusionpolicy,roboagent,rtx,bcz}, which limits their ability to generalize to new objects and function effectively outside of controlled lab environments. In this work, we propose \method{}, which attempts to address this reliance on image representations by directly using points priors as an input to the policy instead of raw images. Through extensive experiments, we observe that such an abstraction helps learn robust policies that generalize across varying scenarios.\looseness=-1

% Our method, \method{}, overcomes this limitation by introducing a more robust point-based state representation, enabling better generalization and performance in real-world settings. \SH{Include approaches that use point-track/flow guidance.}

\subsection{Object-centric Representation Learning}

Object-centric representation learning aims to create structured representations for individual components within a scene, rather than treating the scene as a whole. Common techniques in this area include segmenting scenes into bounding boxes~\cite{deep_object_centric, learning_to_generalize, one_shot, fang2023anygrasp, viola} and estimating object poses~\cite{deep_object_pose_estimation, hope}. While bounding boxes show promise, they share similar limitations with non object-centric image-based models, such as overfitting to specific object instances. Pose estimation, although less prone to overfitting, requires separate models for each object in a task. Another popular method involves using point clouds~\cite{groot, reagentpointcloudregistration}, but their high dimensionality necessitates specialized models, making it difficult to accurately capture spatial relationships. In contrast, \method{} leverages point prescription, eliminating the need to learn a representation because it is predefined by a human. This enables zero-shot generalization to both new objects and new spatial configurations. Similar to our method prior work~\cite{ju2025robo} uses correspondence to identify where to interact with an object, however, this method relies on anygrasp~\cite{fang2023anygrasp}, which is limited to a certain set of objects. Additionally, this method requires learning the affordance of an object, which can introduce errors that are less likely with point prescription.

% However, these methods still suffer from the sample inefficiency typical of RL. Learning from this approach, \method{} applies the key insight that as a robot approaches its goal, the representation should become progressively more similar. This is achieved by constructing a point graph from sparse 3D key points found in the visual representation. \method{} overcomes many of the generalization challenges seen in other methods by utilizing a semantic correspondence model~\cite{dift} and a point tracking model~\cite{cotracker}, both of which have been trained on large vision datasets and have demonstrated strong generalization across diverse scenarios. 
% \SH{Need to change the last part. A comment - isn't point priors more general than something used in RL? I am thinking of works similar to what Lerrel shared.}

% \SH{SSL - BYOL, Moco, MAE, MVP, R3M RoboTap} \\ \SH{Should we focus this section mostly on object-centric representations?}

\section{Background}
\label{sec:background}

\subsection{Behavior Cloning}
 % \LP{Make concise. Maybe cite our recent tutorial} 
 Behavior cloning~\cite{pomerleau1998autonomous,shafiullah2024supervised} aims to learn a behavior policy $\pi^b$ given access to either the expert policy $\pi^e$ or trajectories derived from the expert policy $\mathcal{T}^e$. This work operates in the setting where the agent only has access to observation-based trajectories, i.e. $\mathcal{T}^e \equiv \{(o_t, a_t)_{t=0}^{T}\}_{n=0}^N$. Here $N$ and $T$ denote the number of demonstrations and episode timesteps respectively. We choose this specific setting since obtaining observations and actions from expert or near-expert demonstrators is feasible in real-world settings~\cite{aloha,openteach} and falls in line with recent work in this area~\cite{baku,cbet,vqbet,aloha,diffusionpolicy}.

 % Behavior cloning~\cite{shafiullah2024supervised} aims to learn a behavior policy $\pi^b$ given access to either the expert policy $\pi^e$ or trajectories derived from the expert policy $\mathcal{T}^e$. While there are a multitude of settings with differing levels of access to the expert~\cite{torabi2019recent}, this work operates in the setting where the agent only has access to observation-based trajectories, i.e. $\mathcal{T}^e \equiv \{(o_t, a_t)_{t=0}^{T}\}_{n=0}^N$. Here $N$ and $T$ denote the number of trajectory rollouts and episode timesteps respectively. We choose this specific setting since obtaining observations and actions from expert or near-expert demonstrators is feasible in real-world settings~\cite{aloha,openteach} and falls in line with recent work in this area~\cite{baku,cbet,vqbet,aloha,diffusionpolicy}.

\subsection{Semantic Correspondence}
% Semantic correspondence is a fundamental problem in computer vision that involves identifying matching object parts across multiple images. It is a critical step in many computer vision applications, including 3D reconstruction~\cite{nerf,gs}, motion tracking~\cite{cotracker,pips,pips_plus,tapir}, image registration~\cite{zitova2003image}, and object recognition~\cite{segmentanything}. The goal is to establish reliable correspondences between parts in one image and their counterparts in another image, despite possible variations in viewpoint, lighting, scale, or other factors (e.g., if the original part is the left eye of a dog correspondence models can find the left eye of a cat). Traditionally, correspondence methods have relied on detecting distinctive local features like corners or blobs and matching their descriptors across images~\cite{zitova2003image}. However, this can be challenging in scenes with repetitive patterns, homogeneous regions, or significant appearance changes. More recent approaches leverage deep learning and dense correspondence techniques to improve robustness~\cite{fu2020deep, huang2022learning, asic}. In this work, we use a diffusion-based point correspondence model, DIFT~\cite{dift}, to establish correspondences between a reference image and an observed image (described in Section~\ref{sec:method}). This can be visualized in Figure~\ref{fig:correspondence}.

\begin{figure}[t]
\centering
\includegraphics[width=\linewidth]{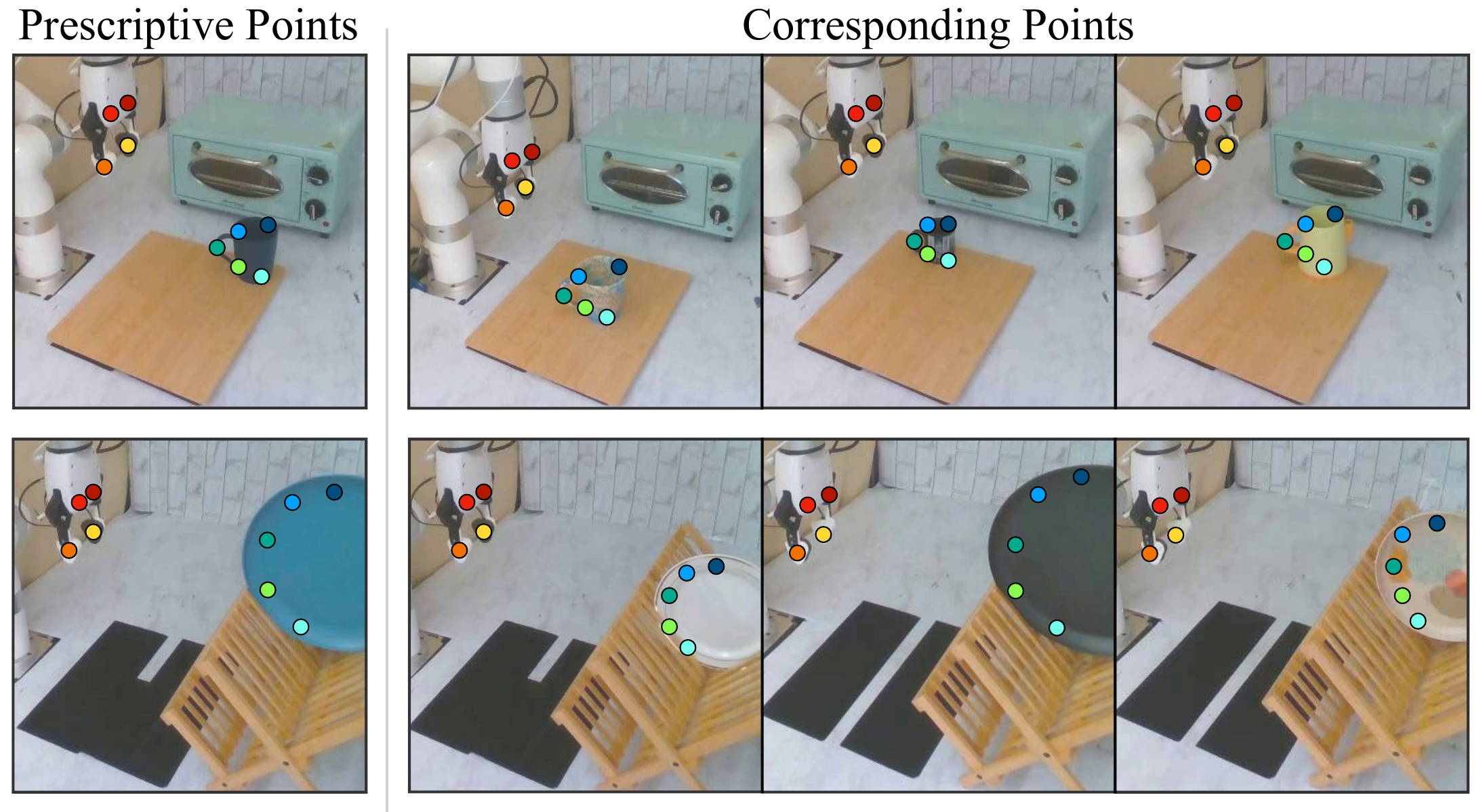}
\vspace{-1em}
\caption{Results of the correspondence model when used on the pick mug and plate off rack tasks. On the left is the frame that is annotated by a human. On the right we show that semantic correspondence~\cite{dift} is able to identify the same points across a variety of instances of each object.}
\vspace{-1em}
\label{fig:correspondence}
\end{figure}

Finding corresponding points across multiple images of the same scene is a well-established problem in computer vision~\cite{sift, zitova2003image}. Correspondence is essential for solving a range of larger challenges, including 3D reconstruction~\cite{nerf,gs}, motion tracking~\cite{cotracker,pips,pips_plus,tapir}, image registration~\cite{zitova2003image}, and object recognition~\cite{segmentanything}. In contrast, semantic correspondence focuses on matching points between a source image and an image of a different scene (e.g., identifying the left eye of a cat in relation to the left eye of a dog). Traditional correspondence methods~\cite{zitova2003image, sift} often struggle with semantic correspondence due to the substantial differences in features between the images. Recent advancements in semantic correspondence utilize deep learning and dense correspondence techniques to enhance robustness~\cite{fu2020deep, huang2022learning, asic} across variations in background, lighting, and camera perspectives. In this work, we adopt a diffusion-based point correspondence model, DIFT~\cite{dift}, to establish correspondences between a reference and an observed image, which is illustrated in Figure~\ref{fig:correspondence}.

\subsection{Point Tracking}

Point tracking across videos is a problem in computer vision, where a set of reference points are given in the first frame of the video, and the task is to track these points across multiple frames of the video sequence. Point tracking has proven crucial for many applications, including motion analysis~\cite{aggarwal1999human}, object tracking~\cite{yilmaz2006object}, and visual odometry~\cite{nister2004visual}. The goal is to establish reliable correspondences between points in one frame and their counterparts in subsequent frames, despite challenges such as changes in illumination, occlusions, and camera motion. While traditional point tracking methods rely on detecting local features in images, more recent advancements leverage deep learning and dense correspondence methods to improve robustness and accuracy~\cite{cotracker, pips, pips_plus}. In this work, we use Co-Tracker~\cite{cotracker} to track a set of reference points defined in the first frame of a robot's trajectory. These points tracked through the entire trajectory are then used to train generalizable robot policies for the  real world. This can be visualized in Figure~\ref{fig:method}.a.

% Although $C$ is able to determine $P^{\text{experiment}}_0$ the computation time of $C$ makes it infeasible for use at every step of the training. Therefore \name uses a point tracking model[CITE HERE], referred to as $T$, to locate the key points from $t=0$ at all other timesteps of an episode. Using $T$ over $C$ is not only useful for total computation time, but also useful because many point tracking methods [CITE HERE] can track points through occlusions the correspondence models would struggle with. We define $T$ as

% \begin{equation}
% \label{eq:tracking}
% P^{\text{experiment}}_t = T(P^{\text{experiment}}_{t - n}, P^{\text{experiment}}_{t - n + 1},\dots, P^{\text{experiment}}_{t - 1}, I^{\text{experiment}}_{t - n}, I^{\text{experiment}}_{t - n + 1},\dots, I^{\text{experiment}}_{t - 1})
% \end{equation}

% The tracking model requires $n$ frames, but \name starts out with information about only one frame. We mediate this by repeating the first frame $n$ times at the start of the tracking. This can be seen in Figure~\ref{fig:tracking}.

% \section{\method{}}
\section{Prescriptive Point Priors for Policies (\method{})}
\label{sec:method}

Given demonstrations for robot manipulation tasks that cover a small set of possible object configurations and types, we seek to learn a generalizable robot policy that is robust to significant environmental variations and applicable to diverse object locations and types. To achieve this, we introduce \method{}, an algorithm that decouples perception and planning to promote generalization. \method{} operates in two phases. First, given a small set of robot demonstrations, the user annotates a single task frame with a set of semantically meaningful points. These reference points are propagated to the rest of the dataset using a combination of semantic correspondence and point tracking. The points obtained are fed into a transformer-based policy model for action prediction. An overview of our method has been provided in Figure~\ref{fig:method}. Below, we describe each component in detail.

% Given a few demonstrations for robot manipulation tasks that cover a small set of possible object configurations and types, we seek to learn a generalizable robot policy that is robust to significant environmental variations and applicable to diverse object locations and types. To achieve this, we introduce \method{}, an algorithm that decouples perception and planning to promote generalization. \method{} operates in three phases. First, given a small set of robot demonstrations, the user annotates one frame per task with a set of object-centric reference points. These reference points are propagated to the rest of the dataset using a combination of semantic correspondence and point tracking. In the second phase, the points in each frame are transformed into a graph-based representation encoding inter-object distances. Finally, these graph representations are fed into a transformer-based policy model for action prediction. Below, we describe these three components in detail,

% \subsection{Data Generation}
\subsection{One-Time Point Prescriptions}
% \LP{Alternate subsection title: One time point prescriptions}
Our method begins by collecting robot demonstrations for a task through robot teleoperation~\cite{openteach}. The user then randomly selects one demonstration and annotates semantically meaningful points on the first frame that are relevant to performing the task, such as points on the robot and the objects being manipulated. This process often requires less than 5 seconds. These user-annotated points serve as priors for the rest of the data generation process. Using an off-the-shelf semantic correspondence model called DIFT~\cite{dift}, we transfer the points on the first frame to the corresponding locations on the first frames of all other demonstrations in the dataset. This allows us to initialize the key points across the entire dataset without additional human effort. For each demonstration, we then employ an off-the-shelf point tracking algorithm, Co-Tracker~\cite{cotracker}, to automatically track the initialized key points through the entire trajectory. In this way, by leveraging existing vision models for correspondence and tracking, we can efficiently compute the key points on every frame in the dataset while requiring the user to only annotate a single frame. This process, visualized in Figure 1, takes advantage of large, internet-scale pre-training of the vision models to generalize to new object instances and scenes without additional training. We prefer point tracking over correspondence at every frame due to its faster inference speed and ability to handle occlusions by continuing to track points. During inference, DIFT is used to mark the corresponding key points on the first frame, followed by Co-Tracker tracking the points during execution. 

\begin{figure}[t]
    \centering
    \includegraphics[width=0.9\linewidth]{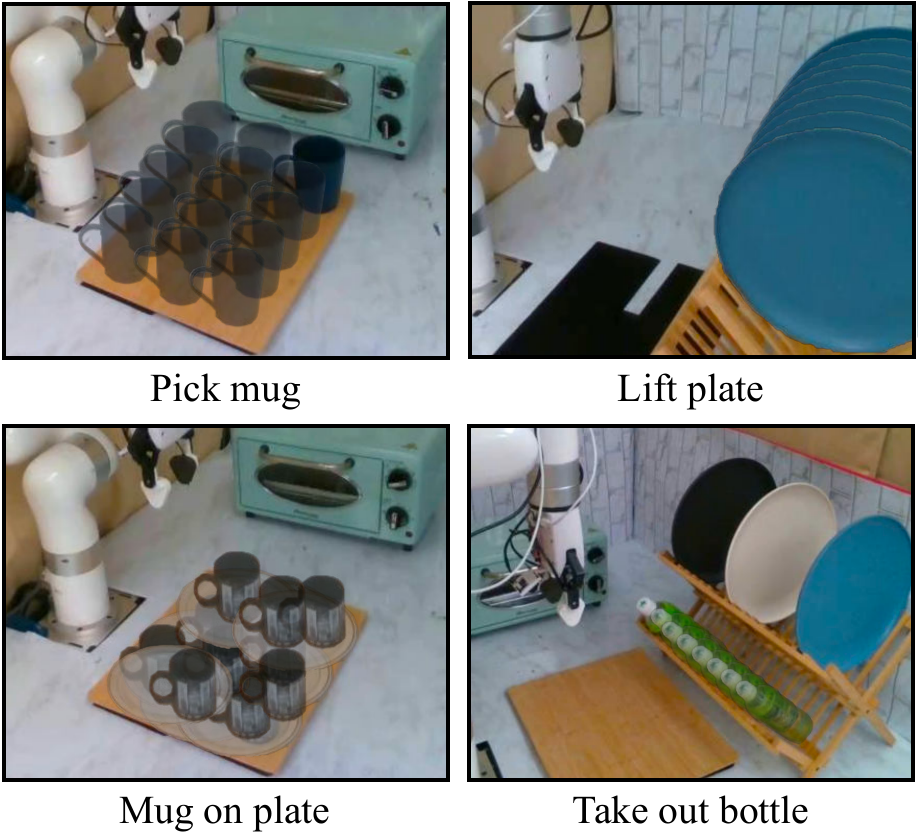}
    \caption{Illustration of spatial variation used in our experiments.}
    \vspace{-1em}
    \label{fig:spatial_variation}
\end{figure}

\subsection{Policy Architecture}
We employ the BAKU~\cite{baku} architecture for policy learning. Instead of feeding in raw images, we use points derived from the previous section as input to the policy. Each 2D image point is first back-projected to 3D using the depth information from the camera. The points are flattened into a single vector in the order in which they are annotated. This point representation is then encoded using a multilayer perceptron (MLP) encoder. Given the noise in real-world depth sensing, we aggregate a history of observation features and feed them as separate tokens into the BAKU causal transformer policy. The policy predicts the action corresponding to each historical token using a deterministic action head and action chunking with exponential temporal averaging~\cite{aloha}.

\section{Experiments}
Our experiments are designed to answer the following questions: $(1)$ How well does \method{} work for policy learning? $(2)$ How well does \method{} work for novel object instances? $(3)$ Can \method{} handle background distractors? $(4)$ How does \method{} perform with estimated depth? $(5)$ Can \method{} be improved with stronger priors?

\subsection{Experimental Setup}
Our experiments are performed on a Ufactory xArm 7 robot with an xArm Gripper in a kitchen environment. The policies are trained with RGB-D images from a third-person camera view and robot proprioception as input. The action space is comprised of the robot end effector pose and the gripper state. We collect a total of 160 demonstrations across 4 real-world tasks with varied object positions and types. The demonstrations are collected using a VR-based teleoperation system~\cite{openteach} at a 30Hz frequency, which are then subsampled to 5Hz. The learned policies are deployed at 5Hz.

% \SH{Add policy learning details - different image sizes, cotracker details (online version), etc.}

% Include details about robot teleoperation, data collection, include collection and deployment frequencies. Include cotracker details (n frames tracked at a time - so we use online version). Image sizes. 

\begin{figure}[t]
    \centering
    \includegraphics[width=0.9\linewidth]{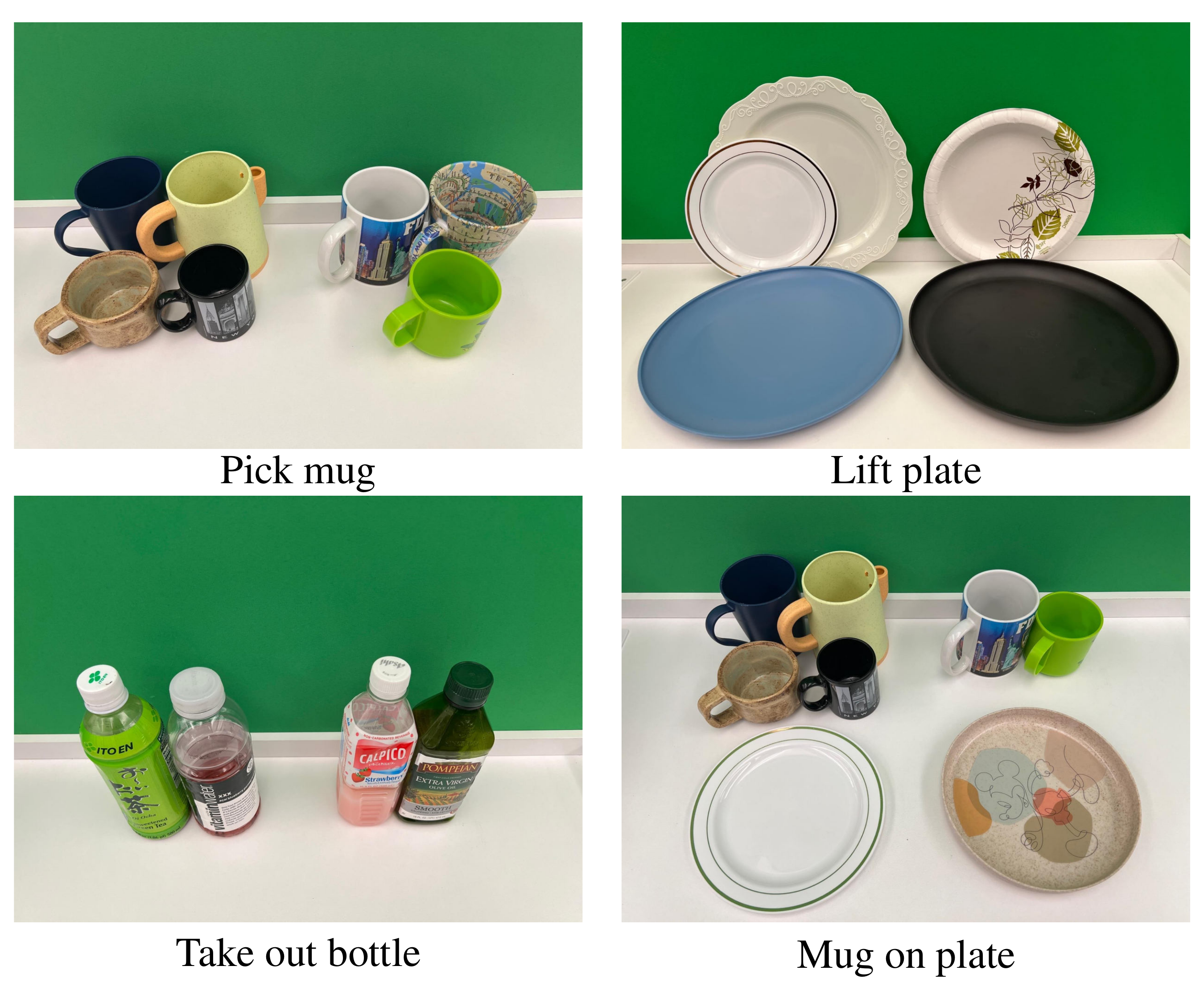}
    \vspace{-0.5em}
    \caption{Illustration of objects used in our experiments. In each image, the left pile depicts the in-domain objects while on the right are novel objects used in our generalization experiments.}
    \vspace{-1.1em}
    \label{fig:train_test_objects}
\end{figure}

\subsection{Task Descriptions}
\begin{figure*}[th]
    \centering
    \includegraphics[width=\linewidth]{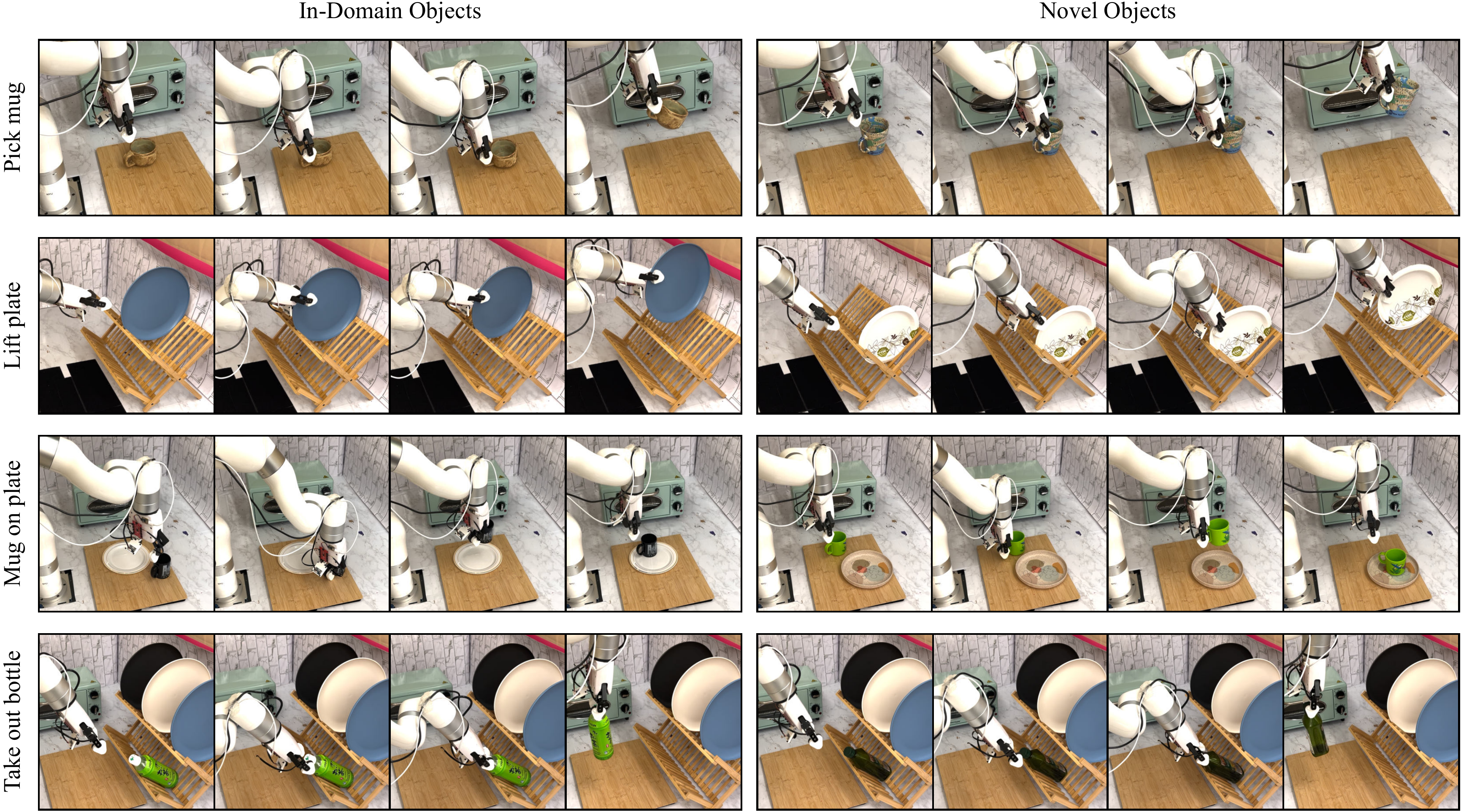}
    \vspace{-1.5em}
    \caption{Real-world rollouts showing \method{}'s  capability on \textbf{(a)} in-domain objects and \textbf{(b)} novel objects.}
    \vspace{-1em}
    \label{fig:rollouts}
\end{figure*}

We experiment with four manipulation tasks that exhibit significant variability in object position, type, and background context. Figure~\ref{fig:rollouts} provides rollouts of the tasks performed in our real-world setup. For each task, we collect expert demonstrations across a variety of object sizes and appearances. We refer to objects and environments seen in our collected data as in-domain. During evaluations, we add novel object instances that are unseen in the training data. The variations in positions and object instances for each task are depicted in Figure~\ref{fig:spatial_variation} and Figure~\ref{fig:train_test_objects} respectively. We provide a brief description of each task below.

\paragraph{Pick mug} The robot arm picks up a mug placed on the kitchen counter. The position of the mug is varied for each evaluation. We collect 15 demonstrations for 4 different mugs, resulting in a total of 60 demonstrations for the task. During evaluation, we introduce 3 novel mugs.

\paragraph{Lift plate} The robot arm lifts a plate placed on the upper level of a rack. We collect 8 demonstrations for 3 different plates, resulting in a total of 24 demonstrations. During evaluation we introduce 2 novel plates.

\paragraph{Mug on plate} The robot arm picks up a mug placed on the kitchen counter and places it on a plate. We collect 15 demonstrations for 4 different mugs placed on the same plate, resulting in a total of 60 demonstrations for the task. During evaluation we use 1 novel mug and 1 novel plate.

\paragraph{Take out bottle} The robot arm takes a bottle out from the lower level of a rack. We collect 8 demonstrations for 2 different bottles, leading to a total of 16 demonstrations. During evaluation we introduce 2 novel bottles.

For all tasks, the xArm is initialized at its home position, while the object locations are varied across trials. During evaluation, the objects are placed in a held-out set of positions to keep comparisons fair across baselines.

\subsection{Baselines} We compare \method{} with three primary baselines. 
% \LP{Better titles needed for baselines. Maybe Full RGB representation}

\paragraph{Full RGB Representation} This method utilizes the BAKU transformer architecture~\cite{baku}, which takes the full RGB image of the scene and robot proprioception as input. 

\paragraph{RGB-D Representation} This is a depth-based extension of BAKU that separately processes the depth image using an encoder and appends a depth token to the policy input for action prediction. 

\paragraph{GROOT~\cite{groot}} GROOT is a transformer-based imitation learning algorithm that constructs an object-centric 3D representation using Segment Anything~\cite{segmentanything} and a clustered point cloud which makes it robust to background distractors and novel objects. We refer the reader to the paper~\cite{groot} for more details about GROOT.

% \paragraph{BAKU~\cite{baku}}: BAKU is a state-of-the-art transformer-based behavior cloning architecture that works with multiple image inputs and robot proprioception.

% \paragraph{BAKU-D}: BAKU-D is a depth-based extension of BAKU which separately processes the depth image using an image encoder and appends a depth token to the policy input for action prediction. 

% \paragraph{GROOT~\cite{groot}}: GROOT is a transformer-based imitation learning algorithm that constructs an object-centric 3D representation using Segment Anything~\cite{segmentanything} and a clustered point cloud which makes it robust to background distractors and novel objects. We refer the reader to the paper~\cite{groot} for more details about GROOT.

\subsection{Considerations for policy learning} 
% Different methods are trained using input modalities suited to their respective approaches. Policies trained with full RGB and RGB-D representation utilize images of size 256x256 pixels directly as their visual inputs. GROOT generates object-centric 3D representations from 480x480 pixel images. \method{} generates a point-based representation from 512x512 pixel images. Our policy uses the point representation instead of raw images, allowing us to use a larger image size without affecting training time. Using large images allows for more precise correspondence in \method{}. For correspondence, \method{} leverages DIFT~\cite{dift}, using the first layer of the hundredth time step with an ensemble size of 4. Point tracking in \method{} is performed by a modified version of online Co-Tracker that enables tracking one frame at a time, rather than in chunks. \method{} and GROOT utilize observation history while the RGB and RGB-D baselines do not~\cite{baku}. Additionally, the RGB and RGB-D baselines incorporate robot proprioception as an input,  while we follow GROOT and do not use robot proprioception as an input to \method{}.

\method{} generates a point-based representation from 512x512 pixel images. For correspondence, \method{} leverages DIFT~\cite{dift}, using the first layer of the hundredth time step with an ensemble size of 4. Point tracking in \method{} is performed by a modified version of online Co-Tracker that enables tracking one frame at a time, rather than in chunks. \method{} and GROOT utilize observation history while the RGB and RGB-D baselines do not~\cite{baku}. Additionally, the RGB and RGB-D baselines incorporate robot proprioception as an input,  while we follow GROOT and do not use robot proprioception as an input to \method{}.

% This larger image size allows for more precise correspondence compared to image-based policies, without affecting training time since the policy only uses the graph representation, not raw images.

% We train \method{} and the baselines using different input modalities suited to each approach. BAKU and BAKU-D were trained on RGB (and depth) images of size $256 \times 256$ pixels, as they directly consume visual inputs. GROOT generates object-centric 3D representations from $480 \times 480$ pixel images. For \method{}, we generate a graph-based representation (as described in Section~\ref{subsec:graph}) on images of size $512 \times 512$ pixels. Our correspondence model uses the first layer of the hundredth time step from DIFT with an ensemble size of 4. \method{} is able to use a large image size compared to image-based policies because the policy is solely trained on our proposed graph-based representation, without any direct use of visual inputs. Therefore using a larger image for this step does not affect the overall training time or final policy size.

% For tracking the points, \method{} uses a modified version of online Co-Tracker~\cite{cotracker} which enables tracking one frame at a time instead of chunks of eight frames used in the original version. Both \method{} and GROOT use a history length of $10$, while the RGB and RGBD models do not use history~\cite{baku}. The RGB and RGBD baselines use robot proprioception as input while we follow GROOT and do not use robot proprioception as an input to \method{}.

\begin{table}[t]
\centering
\caption{In-domain policy performance}
\label{table:same_distribution}
\vspace{-0.05in}
\renewcommand{\tabcolsep}{4pt}
\renewcommand{\arraystretch}{1}
\begin{tabular}{@{}lcccc@{}}
\toprule
\textbf{Method} & \textbf{Pick mug} & \textbf{Lift plate} & \textbf{\begin{tabular}[c]{@{}c@{}}Mug on plate\end{tabular}} & \textbf{\begin{tabular}[c]{@{}c@{}}Take out bottle\end{tabular}} \\ \midrule
RGB~\cite{baku}            & 13/40 & 22/30 & 8/40  & 5/20 \\
RGB-D          & 19/40 & 22/30 & 7/40  & 6/20 \\
GROOT~\cite{groot}           & 7/40  & 7/30  & 0/40  & 0/20 \\ 
\hdashline\noalign{\vskip 0.5ex}
\method{}       & \textbf{39/40} & \textbf{29/30} & \textbf{32/40} & \textbf{13/20} \\ \bottomrule
\end{tabular}
\end{table}

\begin{table}[t]
\centering
\renewcommand{\tabcolsep}{4pt}
\renewcommand{\arraystretch}{1}
\caption{Policy performance on novel object instances}
\label{table:novel_objects}
\vspace{-0.05in}
\begin{tabular}{@{}lcccc@{}}
\toprule
\textbf{Method} & \textbf{Pick mug} & \textbf{Lift plate} & \textbf{\begin{tabular}[c]{@{}c@{}}Mug on plate\end{tabular}} & \textbf{\begin{tabular}[c]{@{}c@{}}Take out bottle\end{tabular}} \\ \midrule
RGB            & 6/30  & 5/20  & 0/20  & 2/20 \\
RGB-D          & 3/30  & 10/20 & 1/20  & 4/20 \\
GROOT           & 4/30  & 5/20  & 0/20 & 0/20 \\ 
\hdashline\noalign{\vskip 0.5ex}
\method{}       & \textbf{29/30} & \textbf{18/20} & \textbf{16/20} & \textbf{10/20}\\ 
\bottomrule
\end{tabular}
\vspace{-1em}
\end{table}

\subsection{How well does \italmethod{} perform for policy learning?}
\label{subsec:spatial_gen}
We first evaluate \method{}'s performance on the in-domain environment configurations using in-domain objects. We conduct 10 trials per object per task resulting in a variable number of total trials per task. The results have been reported in Table~\ref{table:same_distribution}. We observe that \method{} outperforms the strongest baseline on average by $43$\% across our four real-world tasks. It must be noted that we only use 8-15 demonstrations per object per task, which is much smaller than prior works studying spatial generalization in robot policy learning~\cite{umi,aloha,dobbe}. For the full RGB and RGB-D baselines, most failures stem from minor errors, suggesting these policies could be improved with additional demonstrations. In the case of GROOT, having introduced larger variations in both the rotations and spatial locations of objects than the original paper~\cite{groot}, we observe that the policy is unable to generalize. We believe this limitation arises because GROOT normalizes each object-specific point cloud by its centroid, losing information about its position in space. To address this, GROOT adds the positional embedding of the centroid to the processed representation. However, this may not optimally reinforce the positional information. Videos on our website provide examples supporting these hypotheses. 

% \SH{Highlight that performances low because of less demos - only 15 for a large area of variations - have spatial variation plot}

% Additionally, the results reported in GROOT~\cite{groot_cite} were obtained with limited spatial variance, further explaining the performance differences between the two works.

\subsection{How well does \italmethod{} work for novel objects?}
\label{subsec:novel_gen}
Table~\ref{table:novel_objects} compares the performance of \method{} with the baselines when tested on novel objects. These objects can be seen in Figure~\ref{fig:train_test_objects}. We conduct 10 trials for each novel object for each task resulting in a variable number of total trials per task. We observe that \method{}'s visual representation allows it to effectively generalize to novel object instances, outperforming the strongest baseline by $58$\% across all tasks. The RGB and RGB-D baselines exhibit reduced performance on novel objects due to their reliance on visual features learned from the training set. While designed for generalization, GROOT struggles with spatial variations resulting in lower accuracy. These results suggest that \method{}, with its point-based non-image specific representation, is better equipped to generalize to novel objects than prior methods.

\begin{table}[t]
\centering
\renewcommand{\tabcolsep}{4pt}
\renewcommand{\arraystretch}{1}
\caption{Policy performance with background distractors}
\label{table:distractors}
\vspace{-0.05in}
\begin{tabular}{@{}lcccc@{}}
\toprule
\textbf{Method} & \multicolumn{2}{c}{\textbf{Pick mug}}       & \multicolumn{2}{c}{\textbf{Lift plate}} \\ 
\cmidrule(lr){2-3} 
\cmidrule(l){4-5} 
                & In-domain            & Novel object         & In-domain                & Novel object     \\ 
\midrule
RGB            & 0/5  & 0/5  & 4/5  & 0/5 \\
RGB-D          & 1/5  & 0/5  & 2/5  & 0/5 \\
GROOT           & 0/5  & 1/5  & 1/5  & 1/5 \\ 
\hdashline\noalign{\vskip 0.5ex}
\method{}       & \textbf{5/5}  & \textbf{5/5}  & \textbf{5/5}  & \textbf{5/5} \\ 
\bottomrule
\end{tabular}
\vspace{-1em}
\end{table}

\subsection{Can \italmethod{} handle background distractors?}
\label{subsec:distractor}
We evaluate the performance of \method{} in the presence of distractors in the task background. An illustration of the distractors has been included in Figure~\ref{fig:method}(c). We study this on two tasks -  \textit{pick mug} and \textit{lift plate}. For each, we evaluate the performance using 5 trials each on an in-domain object and a novel object. Table~\ref{table:distractors} provides these results. We observe that \method{} outperforms the strongest baseline by 80\%. The image-based baselines exhibit low accuracy due to their reliance on visual features while GROOT struggles with spatial generalization. These results reinforce that \method{}'s point representation, decoupled from raw pixel values, enables policies that are robust to environmental perturbations.

\subsection{How does \italmethod{} without ground truth depth?}
\label{subsec:da}
Given recent advances in monocular depth prediction~\cite{depth_anything_v1,depth_anything_v2}, we investigate the importance of true depth values for the performance of \method{}. To evaluate this, we compare \method{} when using true depth from an RGB-D camera versus predicted depth from an off-the-shelf monocular depth estimation model, Depth Anything 2~\cite{depth_anything_v2}. As shown in Table~\ref{table:da}, we observe that \method{} achieves equivalent performance on two tasks, with one in-domain object and one novel object, regardless of whether true or predicted depth was provided. This is an interesting result, as it implies that \method{} may be applicable to large-scale robot datasets~\cite{rtx,droid} which might not always include depth data.

\begin{table}[t]
\centering
\renewcommand{\tabcolsep}{4pt}
\renewcommand{\arraystretch}{1}
\caption{Effect of camera vs.\ predicted depth on \method{}}
\label{table:da}
\vspace{-0.05in}
\begin{tabular}{@{}lcccc@{}}
\toprule
\multicolumn{1}{@{}l}{\textbf{\method{}}} & \multicolumn{2}{c}{\textbf{Pick mug}}                               & \multicolumn{2}{c}{\textbf{Lift plate}}                         \\ 
\cmidrule(lr){2-3} 
\cmidrule(l){4-5} 
                                    & In-domain & Novel object & In-domain & Novel object \\ \midrule
Camera Depth & 5/5 & 5/5 & 5/5 & 5/5 \\
Depth Anything & 5/5 & 5/5 & 5/5 & 5/5 \\ \bottomrule
\end{tabular}
\vspace{-1em}
\end{table}

\subsection{Can \italmethod{} be improved with stronger priors?}
\label{subsec:stronger_priors}
Prior work has shown that encoding relational structure between inputs can improve policy learning generalization~\cite{kumar2023graph, graph_vi, graph_learning_survey, yin2024offline}. In this section, we investigate whether encoding the spatial relationships between key points as a graph prior could further enhance \method{}'s performance.  Specifically, we represent the key points as a fully connected graph, with edges encoding the 3D distance between each pair of points. Leveraging the annotation order, we flatten this graph into a vector representation encoding the spatial relations between all point pairs. Policies are then trained on this graph-structured input. As shown in Table~\ref{table:stronger_priors}, encoding the key points as a graph prior results in similar performance to directly using the key points as input. While additional structure did not provide clear benefits in this case, future work could explore more sophisticated relational encodings or combining our approach with other structural priors.

\subsection{Can \italmethod{} complete more complex tasks?}
\label{subsec:diff_objects}

\begin{figure*}[th]
    \centering
    \includegraphics[width=\linewidth]{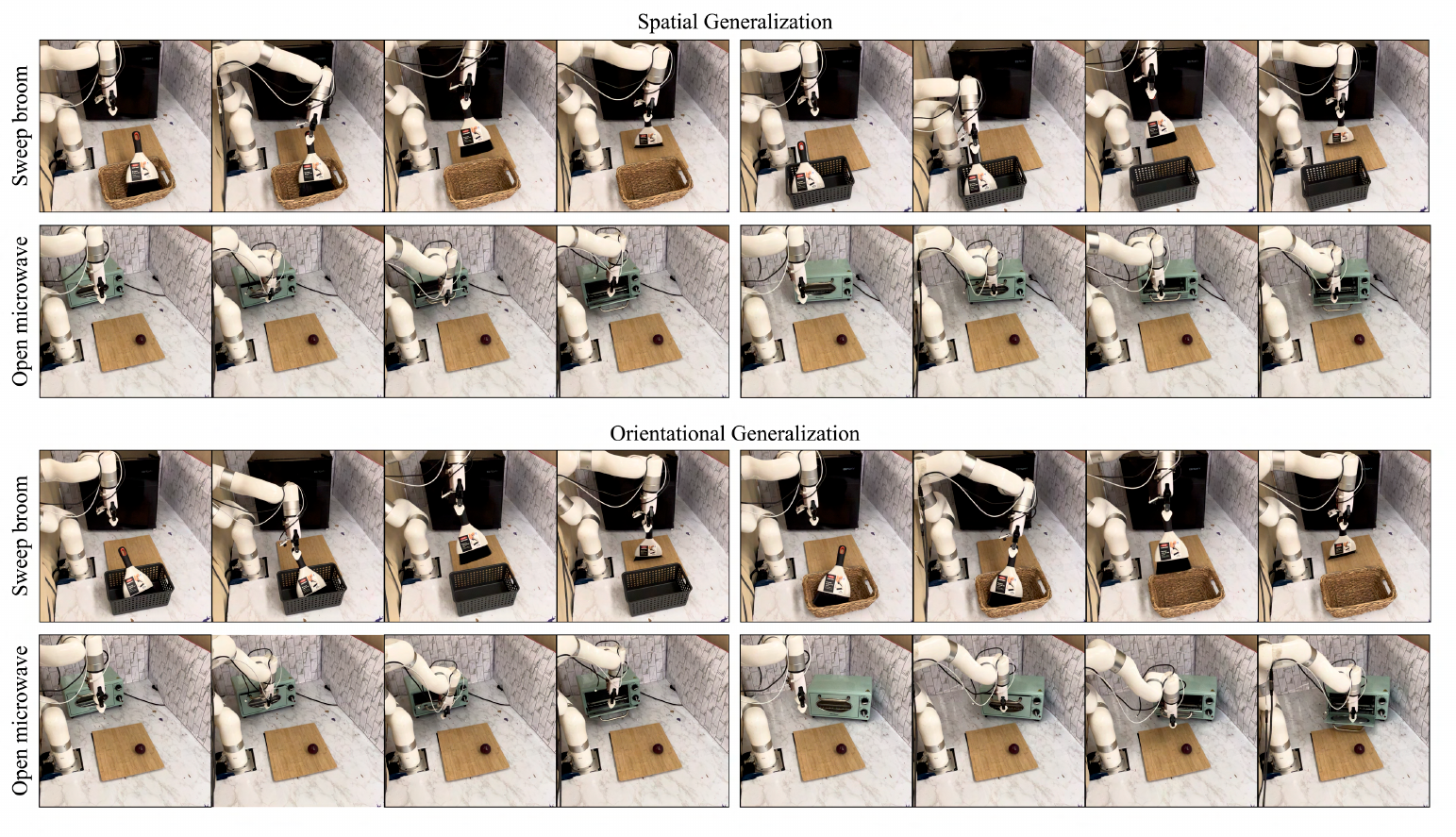}
    \vspace{-1.5em}
    \caption{Rollouts of the two complex tasks. On the top we show that for both of these tasks P3-PO can generalize to the object being in a different location. On the bottom we show that P3-PO can also generalize to different orientations of the object.}
    \vspace{-1em}
    \label{fig:generalization_new_tasks}
\end{figure*}

In this section, we demonstrate that in addition to the tasks shown above, P3-PO excels at tasks that are more complex and dexterous than simple pick-and-place operations. First, we present results for a sweeping task where the robot picks up a broom and sweeps a nearby cutting board (Fig~\ref{fig:generalization_new_tasks}). This task is challenging for two reasons - (1) The broom's handle is rounded, requiring precise handling to prevent slipping. This demonstrates that the point-based context provides sufficient environmental understanding for precise manipulation. (2) The task is long-horizon, consisting of two stages: lifting the broom and sweeping the board. P3-PO is able to understand its place in the sequence and act accordingly.

To evaluate P3-PO's adaptability, we tested the task with two different baskets, showcasing its ability to handle varying vertical and horizontal angles. This is demonstrated in Fig~\ref{fig:generalization_new_tasks} The model achieves an $\mathbf{80\%}$ success rate when trained on 30 demonstrations and evaluated on 10 trials.

% \begin{figure*}[th]
%     \centering
%     \includegraphics[width=\linewidth]{graphics/microwave_figure_compressed.pdf}
%     \vspace{-1.5em}
%     \caption{Rollouts of the open microwave task. On the top we show two successful examples in varying locations with the microwave at different angles. On the bottom we show a failure where the robot opens the microwave, but not far enough for it to stay open when the robot pulls away. \SH{Is this discussed in the text somewhere?}}
%     \vspace{-1em}
%     \label{fig:microwave_task}
% \end{figure*}

Next, we present results for an open-microwave task, which requires the robot to navigate into the thin opening between the handle and the microwave. Despite relying solely on point-based input, the robot is able to succeed with a high level of precision. This task is further challenging due to the wide range of orientations and variations in the microwave's initial placement. The results demonstrate P3-PO's ability to generalize to new locations and accurately interpret object orientations. P3-PO achieves an $\mathbf{80\%}$ success rate on this task when evaluated on 10 trials and trained on 22 expert demonstrations. Variations in microwave locations and task execution are shown in Figure~\ref{fig:generalization_new_tasks}.

\begin{figure*}[th]
    \centering
    \includegraphics[width=\linewidth]{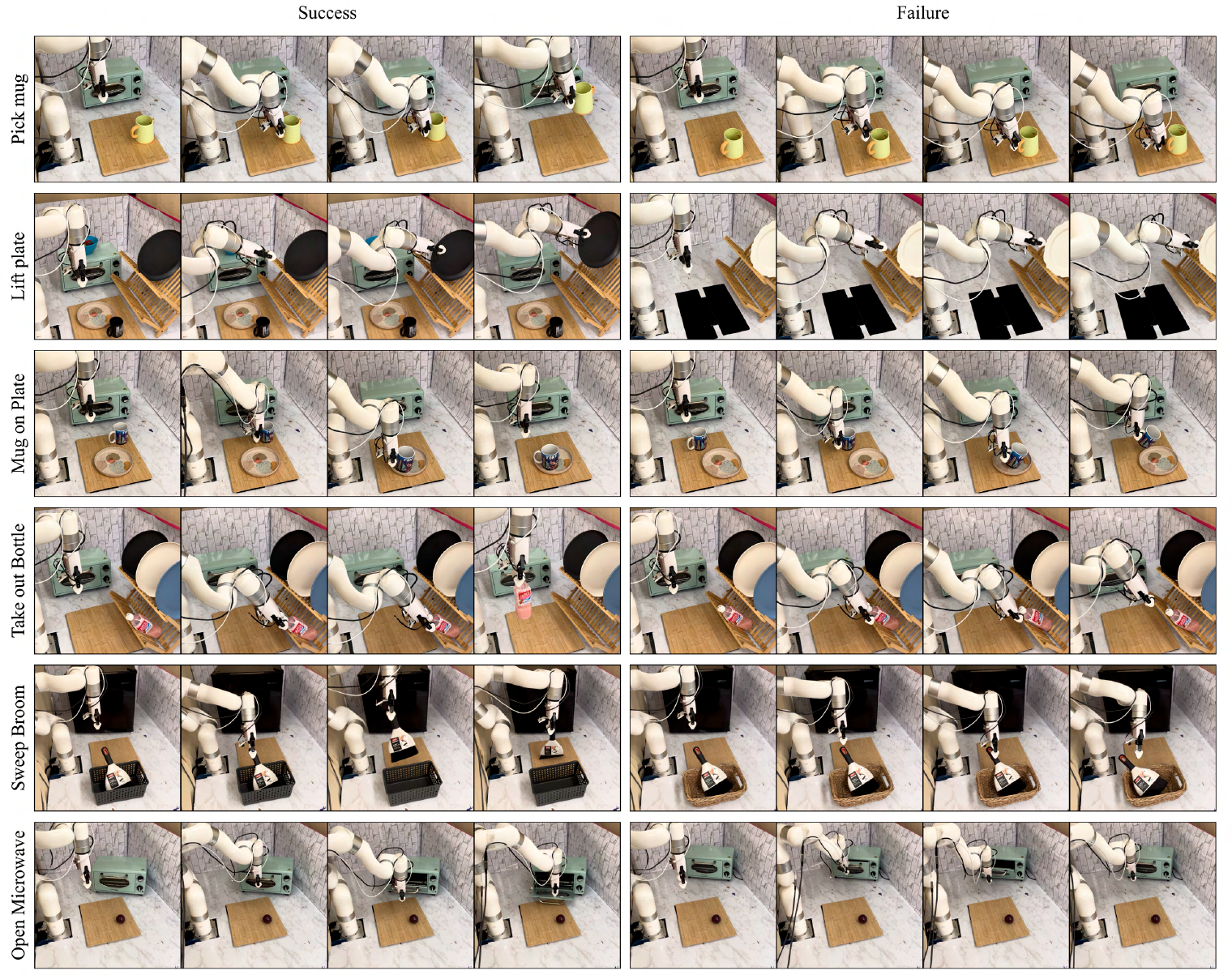}
    \vspace{-1.5em}
    \caption{Demonstrations of success and failure for each task. On the left we show successful demonstrations and on the right demonstrations of episodes that fail.}
    \vspace{-1em}
    \label{fig:failures}
\end{figure*}

\subsection{What do failures look like in \italmethod{}?}
In Figure~\ref{fig:failures}, we present examples of both successful and failed episodes across all six tasks. These examples highlight that while \method{} does not always succeed, its failures come close to achieving the task objectives. For instance, in the "open microwave" task, the failures occur when the robot cannot open the microwave door far enough for the door to remain open. Similarly, tasks like "pick mug" and "lift plate" likely fail due to noise in the depth measurements. Notable these failures are infrequent and we believe they can be addressed in future iterations of this work.

\begin{table}[t]
\centering
\renewcommand{\tabcolsep}{4pt}
\renewcommand{\arraystretch}{1}
\caption{Effect of stronger priors on \method{}}
\label{table:stronger_priors}
\vspace{-0.05in}
\begin{tabular}{@{}lcccc@{}}
\toprule
\multicolumn{1}{@{}l}{\textbf{Input}} & \multicolumn{2}{c}{\textbf{Lift plate}}                               & \multicolumn{2}{c}{\textbf{Take out bottle}}                         \\ 
\cmidrule(lr){2-3} 
\cmidrule(l){4-5} 
                                    & In-domain & Novel object & In-domain & Novel object \\ \midrule
Point & 5/5 & 4/5 & 3/5 & 3/5 \\
Graph & 5/5 & 5/5 & 4/5 & 3/5 \\ \bottomrule
\end{tabular}
\vspace{-1em}
\end{table}

% \subsection{Can \method{} deal with multiple instances of the same object class?}
% \SH{Can we pose this as multi-task behavior from single-task training?}

\section{Conclusion and Limitations}
In this work, we presented Prescriptive Point Priors for Policies (\method), a simple yet effective framework that leverages human-provided semantic key points to enable more robust policy learning. \method{} demonstrates improved generalization to spatial variations, novel objects, and distracting backgrounds compared to prior state-of-the-art methods. We recognize a few limitations in this work: $(a)$ \method{}'s reliance on existing vision models makes it susceptible to their failures. For instance, point tracking failures under occlusion hurt policy performance. However, we believe that continued advances in computer vision will serve to further strengthen performance of \method{}. $(b)$ While point abstractions facilitate better generalization, they lose information about scene context that could be important for navigation amid obstacles or clutter. Future work developing algorithms to retain sparse contextual cues while maintaining \method{}'s object-centric representation may help address this. $(c)$ In this work, we primarily study the single task performance of point prior policies. Extending the framework to multitask learning would be an interesting research direction. Overall, we believe \method{} takes an important step toward developing general, data-efficient robot policies suitable for real-world deployment by grounding them in human point priors.

% In this work, we present Point Priors for Policies (\method), a simple yet effective framework that leverages human-provided point priors to enable more robust policy learning. \method{}  exhibits improved generalization to spatial variations, novel objects, and distracting backgrounds compared to prior state-of-the-art methods.

% SH:
% Limitations: (1) abstracting away all environment info might lead to issues when dealing with cluttered scenes where the robot might hit other objects. 
% (2) We use naive flattening of graph which might not be possible with a variable number of objects in multitask settings. Better architectures such as Graph Nets or Point transformers might be worth looking into

% Future work: (1) Extend this to more complex manipulation tasks / precise tasks. (2) Use online learning to further improve the performance of these policies on novel scenarios. (3) Training such a point-based policy on internet scale data could result in robust robot foundation models.

% ML:
% Checking occlusions - worse keypoints

\section{Acknowledgments}
% This work was partially supported by NSF CAREER Award (\#2238769) to AS. The U.S. Government is authorized to reproduce and distribute reprints for Governmental purposes notwithstanding any copyright annotation thereon. The views and conclusions contained herein are those of the authors and should not be interpreted as necessarily representing the official policies or endorsements, either expressed or implied, of NSF or the U.S. Government.

We would like to thank Raunaq Bhirangi and Venkatesh Pattabiraman for their valuable feedback and discussions. This work was supported by grants from Honda, Google, NSF award 2339096, and ONR awards N00014-21-1-2758 and N00014-22-1-2773. LP is supported by the Packard Fellowship. AS is supported by NSF CAREER Award (\#2238769). The U.S. Government is authorized to reproduce and distribute reprints for Governmental purposes notwithstanding any copyright annotation thereon. The views and conclusions contained herein are those of the authors and should not be interpreted as necessarily representing the official policies or endorsements, either expressed or implied, of NSF or the U.S. Government.

\addtolength{\textheight}{0cm}   % This command serves to balance the column lengths

%%%%%%%%%%%%%%%%%%%%%%%%%%%%%%%%%%%%%%%%%%%%%%%%%%%%%%%%%%%%%%%%%%%%%%%%%%%%%%%%

\bibliographystyle{IEEEtran}
\small
\bibliography{references}

\end{document}